\documentclass[twoside,11pt]{article}
\pdfoutput=1

\usepackage{jmlr2e}

\usepackage{color}
\usepackage{fancyhdr}
\usepackage{microtype}
\usepackage{subfigure}
\usepackage{booktabs} 
\usepackage{enumitem}
\definecolor{Green}{RGB}{10,200,100}

\usepackage[usenames,dvipsnames]{xcolor}
\PassOptionsToPackage{usenames,dvipsnames}{xcolor}
\usepackage{listings}
\usepackage{hyperref}

\definecolor{strings}{rgb}{.624,.251,.259}
\definecolor{keywords}{rgb}{.224,.451,.686}
\definecolor{comment}{rgb}{.322,.451,.322}

\lstdefinelanguage{python}{
  morekeywords={from, import, as, for, in, while, def, return, 
  =, +, -, /, *, lambda},
  keywords=[3]{sample, param, module, Marginal, Posterior, Trace, Poutine, Distribution},
  morecomment=[l]{\#},
  morecomment=[s]{"""}{"""},
  morestring=[b]',
  morestring=[b]",
  alsoletter={<>=-+/*},
  sensitive=true
}

\renewcommand{\texttt}[1]{\lstinline[basicstyle=\fontsize{8pt}{8.25pt}\selectfont\ttfamily]{#1}}

\usepackage[draft=true]{minted}

\jmlrheading{X}{XXXX}{X-X}{06/19}{XX/XX}{paper18a}{authors}  

\ShortHeadings{Pymc-learn: Practical Probabilistic Machine Learning in Python}{Emaasit et al.}
\firstpageno{1}

\begin{document}

\title{Pymc-learn: Practical Probabilistic Machine Learning in Python}

\author{\name Daniel Emaasit \email demaasit@haystax.com \\
       \addr Data Science Team\\
       Haystax Technology\\
       McLean, VA 22102, USA
       }

\editor{XXX}  

\maketitle

\begin{abstract}
  \textit{Pymc-learn} is a Python package providing a variety of state-of-the-art probabilistic models for supervised and unsupervised machine learning. It is inspired by \textit{scikit-learn} and focuses on bringing probabilistic machine learning to non-specialists. It uses a general-purpose high-level language that mimics \textit{scikit-learn}. Emphasis is put on ease of use, productivity, flexibility, performance, documentation, and an API consistent with \textit{scikit-learn}. It depends on \textit{scikit-learn} and \textit{pymc3} and is distributed under the new BSD-3 license, encouraging its use in both academia and industry. Source code, binaries, and documentation are available on \href{http://github.com/pymc-learn}{http://github.com/pymc-learn/pymc-learn}.
\end{abstract}



\begin{figure}[!htb]
\minipage{0.5\textwidth}
\begin{lstlisting}[language=python]
# Linear regression in Pymc-learn
from pmlearn.linear_model \
  import LinearRegression
lr = LinearRegression()
lr.fit(X_train, y_train)
lr.score(X_test, y_test)
lr.predict(X_test)
lr.save("path/to/saved-model")


# Gaussian process regression in Pymc-learn
from pmlearn.gaussian_process \
  import GaussianProcessRegressor()
gpr = GaussianProcessRegressor()
gpr.fit(X_train, y_train)
gpr.score(X_test, y_test)
gpr.predict(X_test)
gpr.save("path/to/saved-model")
\end{lstlisting}
\endminipage 
\hfill
\minipage{0.5\textwidth}
\begin{lstlisting}[language=python]
# Linear regression in Scikit-learn
from sklearn.linear_model \
  import LinearRegression
lr = LinearRegression()
lr.fit(X_train, y_train)
lr.score(X_test, y_test)
lr.predict(X_test)
lr.save("path/to/saved-model")

# Gaussian process regression in Scikit-learn
from sklearn.gaussian_process \
  import GaussianProcessRegressor()
gpr = GaussianProcessRegressor()
gpr.fit(X_train, y_train)
gpr.score(X_test, y_test)
gpr.predict(X_test)
gpr.save("path/to/saved-model")
\end{lstlisting}
\endminipage
\caption{An example comparing \textit{pymc-learn} models and \textit{scikit-learn} estimators: the probabilistic models (\texttt{LinearRegression} and \texttt{GaussianProcessRegressor}) are \texttt{pymc3.Model} objects. The \texttt{fit} method estimates model parameters using either variational inference (\texttt{pymc3.ADVI}) or MCMC (\texttt{pymc3.NUTS}).}
\label{fig:pymc_learn_example}
\end{figure}
\section{Introduction}

Currently, there is a growing need for principled machine learning approaches by non-specialisits in many fields including the pure sciences (e.g. biology, physics, chemistry), the applied sciences (e.g. political science, biostatistics), engineering (e.g. transportation, mechanical), medicine (e.g. medical imaging), the arts (e.g visual art), and software industries. This has lead to increased adoption of probabilistic modeling. This trend is attributed in part to three major factors: (1) the need for transparent models with calibrated quantities of uncertainty, i.e. "\textit{models should know when they don't know}", (2) the ever-increasing number of promising results achieved on a variety of fundamental problems in AI \citep{ghahramani2015probabilistic}, and (3) the emergency of probabilistic programming languages (PPLs) that provide a flexible framework to build richly structured probabilistic models that incorporate domain knowledge. However, usage of PPLs requires a specialized understanding of probability theory, probabilistic graphical modeling, and probabilistic inference. Some PPLs also require a good command of software coding. These requirements make it difficult for non-specialists to adopt and apply probabilistic machine learning to their domain problems. 

\textit{Pymc-learn}\footnote{\url{http://pymc-learn.org/}} seeks to address these challenges by providing state-of-the art implementations of several popular probabilistic machine learning models. \textbf{It is inspired by \textit{scikit-learn}} \citep{pedregosa2011scikit} \textbf{and focuses on bringing probabilistic machine learning to non-specialists}. It puts emphasis on ease of use, productivity, flexibility, performance, documentation and an API consistent with \textit{scikit-learn}. The underlying probabilistic models are built using \textit{pymc3} \citep{salvatier2016probabilistic}.

\section{Design Principles}
The major driving factor in the design of \textit{pymc-learn} was to \textbf{prioritize user experience, especially for non-specialists}. This was achieved by adhering to the following design principles.

\textit{Ease of use}. \textit{Pymc-learn} mimics the syntax of \textit{scikit-learn} -- a popular Python library for machine learning -- which has a consistent \& simple API, and is very user friendly. This makes \textit{pymc-learn} easy to learn and use for first-time users.

\textit{Productivity}. \textit{Scikit-learn} users do not have to completely rewrite their code. Users' code looks almost the same. Users are more productive, allowing them to try more ideas faster. (See Figure \ref{fig:pymc_learn_example} for a comparision).

\textit{Flexibility}. This ease of use does not come at the cost of reduced flexibility. Given that \textit{pymc-learn} integrates with \textit{pymc3}, it enables users to implement anything they could have built in the base language.

\textit{Performance}. \textit{Pymc-learn} uses several generic probabilistic inference algorithms, including the No U-turn Sampler \citep{hoffman2014no}, a variant of Hamiltonian Monte Carlo (HMC).  However, the primary inference algorithm is gradient-based automatic differention variational inference (ADVI) \citep{kucukelbir2017automatic}, which estimates a divergence measure between approximate and true posterior distributions. \textit{Pymc-learn} scales to complex, high-dimensional models thanks to GPU-accelerated tensor math and reverse-mode automatic differentiation via Theano \citep{theano2016},
and it scales to large datasets thanks to estimates computed over mini-batches of data in ADVI.

\section{Project Openness and Development}  

Source code for \textit{pymc-learn} is freely available under the new BSD-3 license and developed by the authors and a community of open-source contributors  
at \url{https://github.com/pymc-learn}. Documentation, examples, and a discussion forum
are hosted online at \url{https://pymc-learn.org}.
A comprehensive test suite is run automatically by a continuous integration
service before code is merged into the main codebase to maintain a high level of project quality and usability.

\section{Illustration}
Built distributions of \textit{pymc-learn} are available for download from PyPi. Source code is available on GitHub. For illustration purposes, the following sections describe how \textit{pymc-learn} can be used in a workflow that mimics \textit{scikit-learn}.

\begin{figure}[!htb]
\centering
\caption{Sample data}
\label{fig:sample-data}
\includegraphics[width=0.7\textwidth]{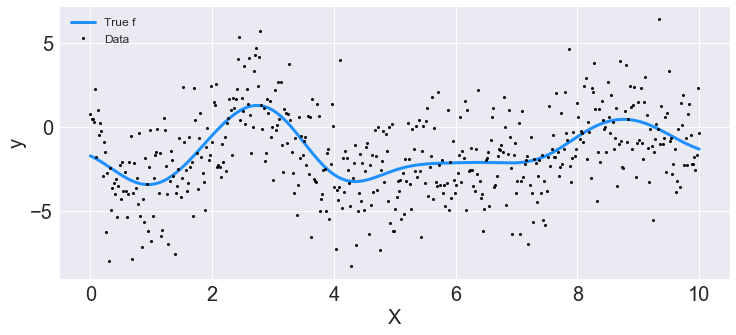}
\end{figure}

Install \textit{pymc-learn} from PyPi:

\begin{minted}{bash}
$ pip install pymc-learn
\end{minted}

Or from source:

\begin{minted}{bash}
$ pip install git+https://github.com/pymc-learn/pymc-learn
\end{minted}

Consider that some data has been imported into a Python environment as shown in Figure \ref{fig:sample-data}. 

\begin{minted}{python}
from sklearn.model_selection import train_test_split
X_train, X_test, y_train, y_test = train_test_split(X, y, test_size=0.3)
\end{minted}


\subsection{Instantiate a model}

Instantiate a model with basic default parameters. For instance, a Gaussian process model uses the squared exponential kernel as the covariance function with default priors for the hyperparameters.

\begin{minted}{python}
# Regression using a Gaussian process model
from pmlearn.gaussian_process import GaussianProcessRegressor
model = GaussianProcessRegressor()
\end{minted}

Methods such as fit, score, predict, save and load are available just like with a \textit{scikit-learn} model.

\subsection{Perform inference}

\begin{minted}{python}
# Estimate using the default ADVI algorithm
model.fit(X_train, y_train)
\end{minted}

\subsection{Score the trained model}

\begin{minted}{python}
# Estimate prediction accurary
model.score(X_test, y_test)
\end{minted}

\subsection{Use the trained model for prediction}

\begin{minted}{python}
# Predict on new data
y_predict = model.predict(X_test)
\end{minted}

\subsection{Save the trained model}

\begin{minted}{python}
# Save for use later in production
model.save('path/to/saved/models')
\end{minted}

\subsection{Load the saved model}

\begin{minted}{python}
# Load the saved model to use in production
model_new = GaussianProcessRegressor()
model_new.load('path/to/saved/models')
model_new.score(X_test, y_test)
\end{minted}

Given that \textit{pymc-learn} is built on top of \textit{pymc3}, common Bayesian workflow methods for diagnozing convengence (such as visualizing traceplots) and critizing results (such as summary tables) are available.

\subsection{Diagnose convergence}

\begin{minted}{python}
# Diagnose convergence using Evidence Lower Bound
model.plot_elbo()
\end{minted}

\begin{figure}[!htb]
\centering
\caption{ELBO plot}
\label{fig:elbo-plot}
\includegraphics[width=0.7\textwidth]{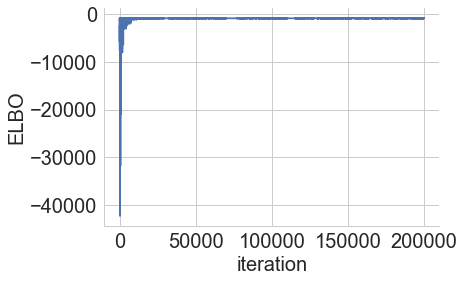}
\end{figure}

\subsection{Visualize traceplots}

\begin{minted}{python}
import arviz as az
az.plot_trace(model.trace);
\end{minted}

\begin{figure}[!htb]
\centering
\caption{Traceplot}
\label{fig:traceplot}
\includegraphics[width=0.8\textwidth]{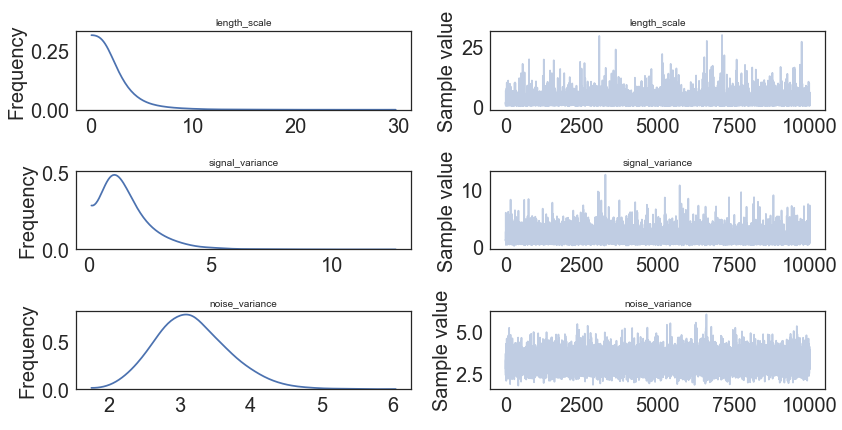}
\end{figure}

\section{Conclusion}
\textit{Pymc-learn} exposes a wide variety of probabilistic machine learning models for both supervised and unsupervised learning. It is inspired by \textit{scikit-learn} with a focus on non-specialists. Future work includes adding more probabilistic models including hidden markov models, Bayesian neural networks, and many others.

\acks{We would like to acknowledge the \textit{scikit-learn}, \textit{pymc3} and \textit{pymc3-models} communities for open-sourcing their respective Python packages.}

\bibliography{paper}

\end{document}